\documentclass[10pt,twocolumn,letterpaper]{article}

\usepackage{cvpr}
\usepackage{times}
\usepackage{epsfig}
\usepackage{graphicx}
\usepackage{amsmath}
\usepackage{amssymb}

\usepackage{hyperref}

\cvprfinalcopy % *** Uncomment this line for the final submission

\def\cvprPaperID{****} % *** Enter the CVPR Paper ID here

\ifcvprfinal\fi
\begin{document}

%%%%%%%%% TITLE
\title{In the sight of my wearable camera: Classifying my visual experience}

\author{Alessandro Perina \\
Microsoft Research\\
One Microsoft Way, Redmond \\
\url{alperina@microsoft.com} \\
\and Nebojsa Jojic \\
Microsoft Research\\
One Microsoft Way, Redmond \\
\url{jojic@microsoft.com}
}

\maketitle
% \thispagestyle{empty}

%%%%%%%%% ABSTRACT
\section{Introduction}

We introduce and we analyze a new dataset which resembles the input to biological vision systems much more than most previously published ones. Our analysis leaded to several important conclusions. First, it is possible to disambiguate over dozens of visual scenes (locations) encountered over the course of several weeks of a human life with accuracy of over 80\%, and this opens up possibility for numerous novel vision applications, from early detection of dementia to everyday use of wearable camera streams for automatic reminders, and visual stream exchange. Second, our experimental results indicate that, generative models such as Latent Dirichlet Allocation \cite{FFPer} or Counting Grids \cite{CG}, are more suitable to such types of data, as they are more robust to overtraining and comfortable with images at low resolution, blurred and characterized by relatively random clutter and a mix of objects.

\section{Data Acquisition}

To gather the data, a subject wore a SenseCam\footnote{\url{http://viconrevue.com/}} during all waking hours for three weeks. The camera was rarely turned off, except during potentially sensitive moments. The SenseCam snapshots\footnote{\url{http://profs.sci.univr.it/~perina/sensecam.htm}} are automatically triggered by sudden changes in the visual field, or by default every 45s. On average the snapshots were taken every 20s or so. This translated into $\sim$2k images a day with a resolution of $640\times480$, for a total of 43522 images. \\

\begin{figure*}[t]
\centering
\includegraphics[width=1\textwidth]{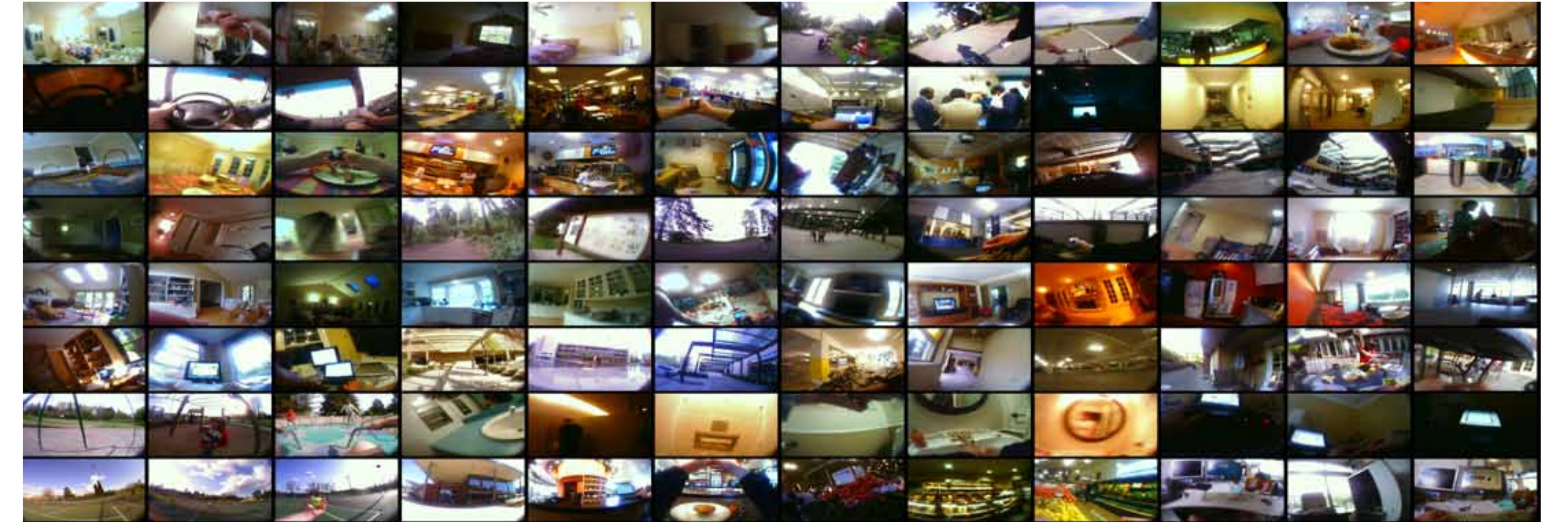}
\caption{From top left to bottom right, we show three images from each class of \emph{SenseCam-32}. The 32 classes are: {\scriptsize \emph{Bathroom Home, Bedroom, Biking, Cafeteria, Car, Classroom, Conference Room, Corridors Work, Dining Room, Bakery, Garage, Atrium, Entry, Hiking Trail, Ice Palace, Kids Bedroom, Game Room, Kitchen, Living Room, Lounge, Home Office, Campus, Parking Work, Patio, Playground, Restroom Work, Small Bathroom Home, Small Home Office, Tennis Court, Food Court, Grocery Store, Work Office}}.}
\label{fig:exp2}
\end{figure*}

We selected a random 10\% of the data, where we found that the recurrent types of scenes fell into 32 classes. About 15\% of images in this random selection belong to spurious types of scenes with only one or two examples. \\
With the help of the original subject, we started from a few examples of each of the 32 recurring scenes and then we manually labeled the rest of the selected frames. In case of some classes, this procedure did not yield enough images for proper training and testing, and for these classes we looked at the whole dataset again and extracted more examples of each of these classes for both testing and training. This process yielded to a total of 3959 labeled images. Some images for each class are shown in Fig.\ref{fig:exp2}. \\

\begin{figure*}[t]
\centering
\includegraphics[width=1\textwidth]{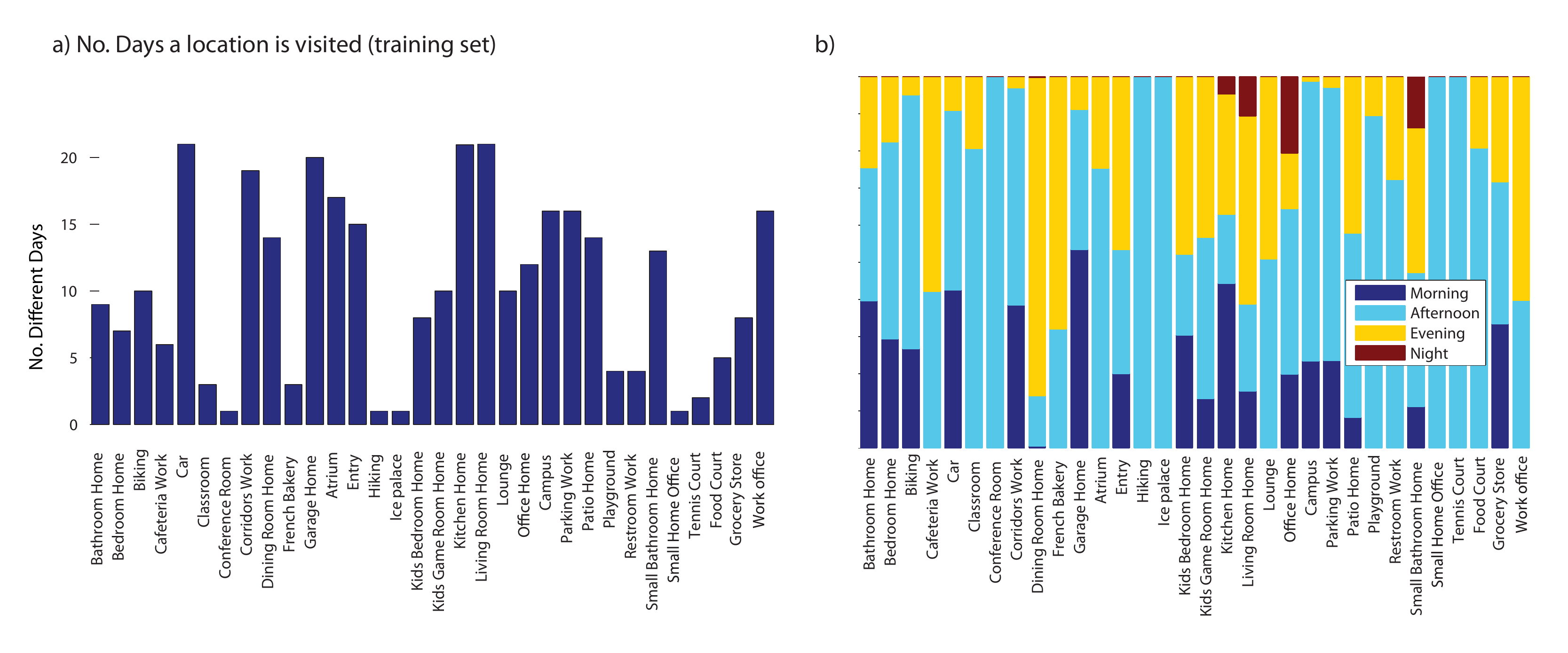}
\caption{Statistics on SenseCam Dataset}
\label{fig:sense1}
\end{figure*}

During acquisition, each image is time-tagged, enabling us to illustrate in Fig.\ref{fig:sense1}a the number of different days that each class was seen, while in Fig.\ref{fig:sense1}b we show the distribution over time of day (morning/afternoon/evening/night) for each class. \\

We also labeled the images coming from two whole days to test if the timestamp information can help recognition. The number of images of each day is, respectively 2043 and 1703. \\
To download the dataset, please send an email to the authors\footnote{\url{http://profs.sci.univr.it/~perina/download/SenseCam_ECCV.zip}}.

\section{Experiments}

In the following experiments, we used SIFT features \cite{sift}, extracted from 16$\times$16 patches spaced 8 pixels apart, clustered in Z=200 visual words or gist descriptors \cite{OT}, extracted on 4 scales with 8 orientations per scale.
We compared the performances of generative, discriminative and epitome-like methods. As discriminative methods we employed a Support Vector Machine with linear kernel on gist \cite{OT} and on quantized sift histograms and the Spatial Pyramid Kernel (SPK, 3 Levels) \cite{spk}. As generative models we considered Fei-Fei's semi-supervised Latent Dirichlet Allocation (LDA) \cite{FFPer}, Counting Grids (CG) \cite{CG}, and a mixture of Dirichlet distributions over the quantized sift histograms. The last approach is similar in spirit to \cite{torralbaICCV} where a mixture of gaussians over gist descriptors is learnt for each class. Finally we also tried epitome-like approaches: Structural Epitome \cite{SE}, Epitomic Location Recognition\footnote{We only used SIFT} \cite{FE} and FFT's Counting Grids \cite{CGFFT} (an hybrid between epitomes and Counting Grids).

\subsection{Scene Classification}
In this section we provide some baseline on the scene/place classification task. Since some categories have much more images than others, we used 15 images per class for training and at most 15 images of each class as test set. We repeated the experiments for 5 times, averaging the results. Classification accuracies are shown in Tab. \ref{tab:vidd}, where we report the best result obtained by each method. \\
Generative classifiers are built learning a model for each class with the training data, and assigning to a test sample the class that produces higher likelihood. As visible in Tab.\ref{tab:vidd} they reach the best results, moreover their performance doesn't vary much for a ``reasonable'' choice of the parameter setting (Topic number of LDA, Capacity for Counting Grids). The poor accuracies of discriminative methods \cite{spk,OT}, are clearly do to overtraining but with this type of datasets we must expect scarce labeled data.
Interestingly, the methods based on pixelwise comparisons fail on this extended dataset as they cannot capture well the geometric transformations.
Being a hybrid between epitomes and counting grids, \cite{CGFFT} reaches decent results, but it also finds it hard to generalize with so little labeled data.
\begin{table}
 \caption{\textbf{Scene Classification Results}. We reported the best result obtained.}
 \label{tab:vidd}
  \begin{center}
\begin{tabular}{|ccc|}
  \multicolumn{3}{c}{\emph{Discriminative Methods}}   \\
  Gist + SVM \cite{OT} & SIFT + SVM & SPK \cite{spk}  \\
 \hline
38,10\% & 49,94\% & 52,76\% \\

 \multicolumn{3}{c}{\emph{Generative Methods}} \\
 LDA \cite{FFPer} & Dirichlet Mixt. \cite{torralbaICCV} & CGs \cite{CG} \\
  \hline
\textbf{58,05}\% & 49,19\% & 54,43\%  \\

 \multicolumn{3}{c}{\emph{Epitome-like Methods}} \\
 Stel Ep. \cite{SE} & CG-fft \cite{CGFFT} & Feature Ep. \cite{FE} \\
  \hline
 15,02\% & 39,40\% & 21,32\%

\end{tabular}
  \end{center}
\end{table}

\subsection{Where was I?}

As second test we asked how many places we could correctly guess using \emph{i)} each of the generative models considered in the previous section, and \emph{ii)} an HMM over possible classes $c_t$, capturing transition constraints such as living room - kitchen, or office - corridor - atrium, etc. \\
We considered labeled images from two days. During each day, the camera bearer visited roughly 20 of the labeled locations and the two days share only 12 locations. Nevertheless we trained models with all the 32 classes as a-priori we cannot know the locations visited during a day. Our goal is to compute the place posterior probabilities at the instant $t$, given all the previous images $P(c_t = k | x_{1:t} )$. We used the forward-backwards procedure to recursively compute it, in formulae:
\begin{eqnarray}
P(c_t = k | x_{1:t} ) \propto  p(x_t|c_t=k)\cdot P(c_t=k|x_{1:t-1}) = \;\;\;\;\;\;\;\;\;\; \nonumber \\
\nonumber \\
\; p(x_t | c_t=k )\cdot \sum_c P( c_{t}=k | c_{t-1} = c)\cdot P(c_{t-1} | x_{1:t-1}) \nonumber
\label{eq:fb}
\end{eqnarray}
We fixed HMM's observation loglikelihood (e.g., $p(x_t|c_t=k)$) to the negative free energy of the generative model in hand, while we used EM estimate the transition matrix $A_{k|c} = P( c_{t} =k | c_{t-1} = c)$, the place prior $\pi_k = P(c_1 =k)$ and the place posteriors an unsupervised way, simply fitting the likelihood to the day's images. We used a strong dirichlet prior over self transitions to favor the stay in the same state/location. Finally since the observation likelihood terms, often dominate the effects of the transition
prior, we adopt the standard solution of re-scaling the likelihood terms. \\
This approach is very similar to \cite{torralbaICCV}, therefore the reference provides a natural point of a comparison. Besides a similar use of the HMM the idea of \cite{torralbaICCV} is to learn a mixture of model for each class to eventually compute the observation likelihood.  \\
We used at most 30 images per class to learn the models. Results are reported in Tab.\ref{tab:vidd2}: as expected, the recognition accuracy rises respectively when we ``turn on'' the HMM (Eq. \ref{eq:fb}).
For sake of completeness we have also implemented the original method of \cite{torralbaICCV} using their descriptors from the whole images and within the four sectors. Their performances were lower ( below 50\%).

\begin{table}
 \caption{\textbf{Where was I?}}
 \label{tab:vidd2}
  \begin{center}
\begin{tabular}{|ccc|}
 \multicolumn{3}{c}{\emph{``HMM off''}} \\
 LDA \cite{FFPer} & Dirichlet Mixt. $\sim$\cite{torralbaICCV} & CGs \cite{CG} \\
  \hline
62.21\% & 54.68\% & \textbf{66.76}\%  \\

 \multicolumn{3}{c}{\emph{``HMM on''}} \\
 LDA \cite{FFPer} & Dirichlet Mixt. $\sim$\cite{torralbaICCV} & CGs \cite{CG} \\
  \hline
76.80\% & 70.37\% & \textbf{81.21}\%
\end{tabular}
  \end{center}
\end{table}

\section{Conclusion}

In this extended abstract we presented a large dataset which differently from others
is totally natural as it represents the visual input of a subject. Using our labels,
it would be easier to analyze the full data collected in \cite{SE}. We also showed how
temporal information can be used to reach compelling accuracies on classification
of all the locations a subject visits during a day.

\end{document}